%% file: AnonymousSubmission2027.tex
\documentclass[letterpaper]{article} % DO NOT CHANGE THIS
\usepackage{aaai2027}  % DO NOT CHANGE THIS
\usepackage[hyphens]{url}  % DO NOT CHANGE THIS
\usepackage{graphicx} % DO NOT CHANGE THIS
\usepackage{natbib}  % DO NOT CHANGE THIS AND DO NOT ADD ANY OPTIONS TO IT
\usepackage{caption} % DO NOT CHANGE THIS AND DO NOT ADD ANY OPTIONS TO IT
\usepackage{algorithm}
\usepackage{algorithmic}

\usepackage{newfloat}
\usepackage{listings}
\DeclareCaptionStyle{ruled}{labelfont=normalfont,labelsep=colon,strut=off} % DO NOT CHANGE THIS
\floatstyle{ruled}
\newfloat{listing}{tb}{lst}{}
\floatname{listing}{Listing}

\usepackage{booktabs}

\usepackage{bibentry}
\usepackage{amssymb}
\usepackage{marvosym}
\usepackage{booktabs}
\usepackage{tabularx}
\usepackage{bbold}
\usepackage{amsmath}
\usepackage{multirow}
\usepackage{subcaption} 
\usepackage[table]{xcolor}  % 支持表格中的颜色
\definecolor{waymolgray}{HTML}{F0F0F0} 
\usepackage{array}    % 支持列对齐控制

\usepackage{enumitem}

\title{Reflect-VLA: Internalizing Reflective Guidance via Dual-Path On-Policy Self-Distillation for Autonomous Driving}
\author {
    Yuechen Luo\textsuperscript{\rm 1}\equalcontrib, He Chong\textsuperscript{\rm 1}\equalcontrib, Fang Li\textsuperscript{\rm 2}\equalcontrib, Shaoqing Xu\textsuperscript{\rm 2}\equalcontrib\projectleader, Hexin Zhang\textsuperscript{\rm 3}, Ruiqi Zhang\textsuperscript{\rm 1}, \\ Ziying Song\textsuperscript{\rm 4}, Lei Yang\textsuperscript{\rm 5}, Zhi-xin Yang\textsuperscript{\rm 2}\corresponding, Fuxi Wen\textsuperscript{\rm 1}\corresponding
}
\affiliations {
    \textsuperscript{\rm 1}Tsinghua University \textsuperscript{\rm 2}University of Macau
    \textsuperscript{\rm 3}University of Science and Technology of China\\
    \textsuperscript{\rm 4}Beijing Jiaotong University
    \textsuperscript{\rm 5}Nanyang Technological University\\
    https://github.com/luo-yc17/Reflect-VLA
}
\newcommand{\method}{Reflect-VLA}

\begin{document}

\maketitle

\input{sec/0_abstract}
\input{sec/1_intro}
\input{sec/2_related}
\input{sec/3_method}
\input{sec/4_exp}

\input{sec/5_conclusion}

\clearpage
\bibliography{aaai2027}

\clearpage

\clearpage
\input{sec/appendix}

\end{document}

%% file: sec/0_abstract.tex
\begin{abstract}
Reinforcement Learning (RL) substantially improves planning performance in autonomous-driving Vision-Language-Action (VLA) models. However, existing RL methods rely on scalar trajectory rewards that measure overall quality but fail to identify specific planning errors or provide actionable correction signals. In long-tail scenarios, errors in scene understanding, behavior selection, and trajectory execution are compressed into a single score. We observe that Reflective Guidance turns trajectory evaluation into explicit diagnoses and corrective instructions, substantially improving planning accuracy. Yet such guidance requires an initial prediction and its evaluation results, making it unavailable for direct single-pass inference. We propose \textbf{\method}, which treats Reflective Guidance as training-time privileged information and internalizes its planning benefits through \textbf{Dual-Path On-Policy Self-Distillation (Dual-OPSD)}. Unlike scalar-reward optimization, Dual-OPSD uses the driving score only to route student-generated trajectories. It distills dense token-level corrective distributions from the teacher for risky trajectories while applying self-imitation to preserve reliable ones. To provide reliable corrective distributions on student rollouts, the teacher is progressively constructed through Driving-Knowledge Pretraining, Planning-and-Refinement SFT, and Self-Refinement Rejection Sampling. At inference time, \method~generates trajectories in a single pass without explicit guidance or iterative refinement. Experiments on NAVSIM v1/v2 and Bench2Drive demonstrate competitive planning performance while reducing training time by over 90\% compared with RL.
\end{abstract}

%% file: sec/1_intro.tex
\section{Introduction}

\begin{figure}[t!]
    \centering
    \includegraphics[width=\linewidth]{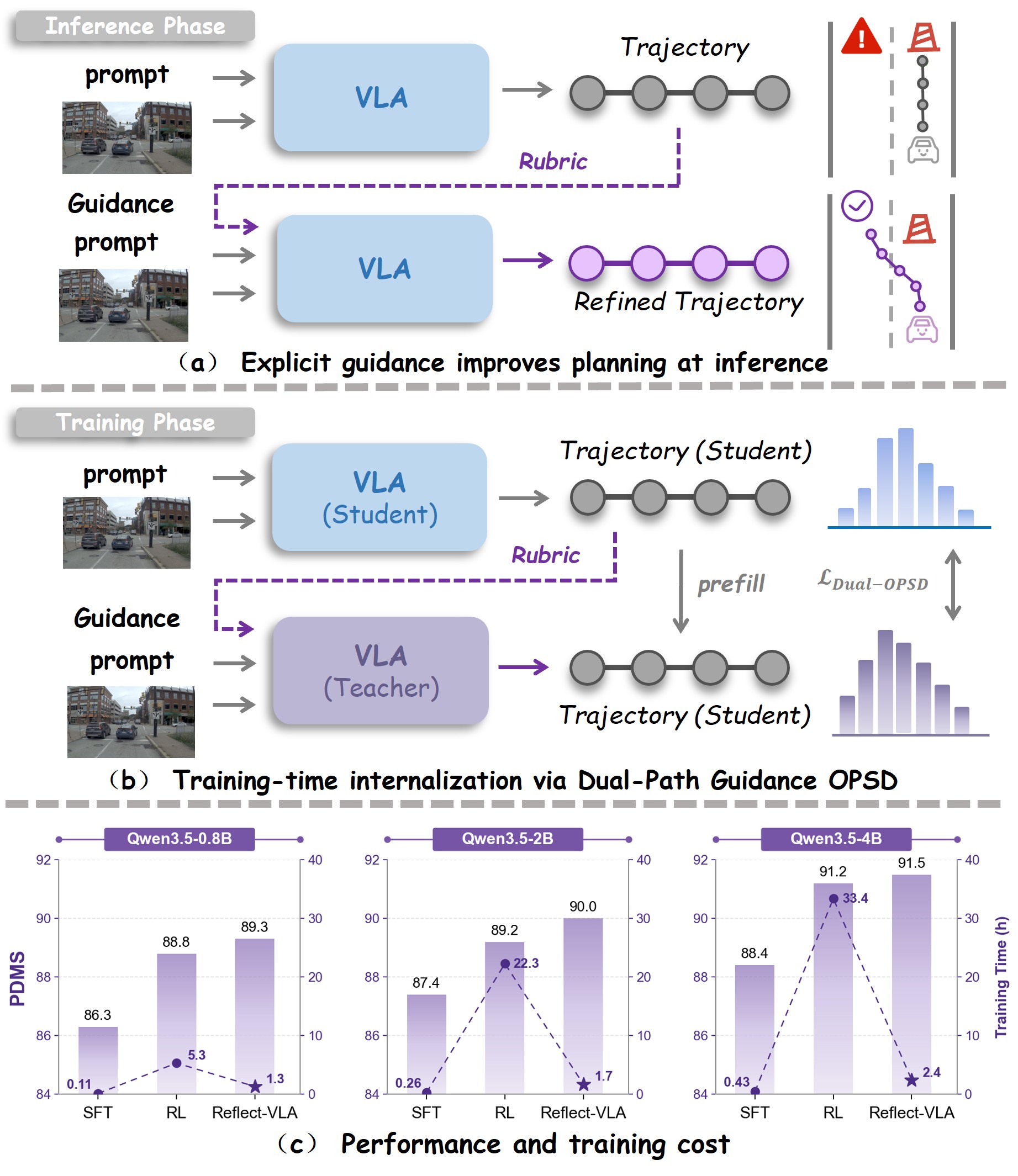}
    \caption{\textbf{From Reflective Guidance to Internalized Planning.}
    (a) Reflective Guidance corrects an unsafe initial trajectory during iterative inference, revealing substantial planning gains.
    (b) Reflect-VLA transfers these gains through Dual-OPSD by aligning teacher and student token distributions along student-generated trajectories.
    (c) Across different model scales, Reflect-VLA improves planning performance over SFT while requiring substantially less training time than RL, enabling guidance-free single-pass planning.}
    \label{fig:intro}
    \vspace{-1em}
\end{figure}

Autonomous driving is shifting from conventional modular pipelines toward end-to-end Vision-Language-Action (VLA) systems~\cite{vlasurvey}. Unlike modular systems~\cite{hu2023planning,jiang2023vad,e2esurvey} that coordinate perception, prediction, and planning through handcrafted interfaces, VLA models integrate visual observations, language instructions and action generation within a unified architecture. Recent driving VLA methods commonly adopt a multi-stage training paradigm~\cite{li2025recogdrive,zhou2025autovla,adathinkdrive,luo2026last,lu2026xiaomi}. They first acquire domain knowledge from large-scale driving question-answering data, then learn trajectory generation through supervised fine-tuning, and finally improve the policy using reinforcement learning or planning-oriented objectives.

Despite progress, VLA models struggle to learn from planning failures, especially in complex and long-tail scenarios. A scalar planning score can identify a poor trajectory but cannot reveal whether the error arises from scene understanding, behavior selection, or trajectory execution, nor indicate how to correct it. Trajectory-level rewards collapse these heterogeneous failure modes into a single value, leaving the model without a direction for improvement. ELF-VLA~\cite{luo2026unleashing} alleviates this problem by introducing structured failure information, showing that detailed feedback can guide the correction of failed trajectories. However, how to convert feedback-assisted correction into an intrinsic single-pass planning capability remains underexplored.

We begin with a key observation: \textbf{Reflective Guidance provides substantial planning gains}. As illustrated in Fig.~\ref{fig:intro}, a VLA model may produce an incorrect trajectory during direct inference, yet generate a markedly better trajectory when provided with evaluation-derived diagnoses and corrective suggestions~\cite{yang2026judge,he2026drivever,chen2026vilta}. Constructing the guidance requires both an initial trajectory and its evaluation results, whereas an autonomous driving system must produce an executable plan before acting. The central challenge is therefore not to perform iterative refinement at inference time, but to transfer the planning preferences induced by Reflective Guidance into the model's direct trajectory-generation policy.

To this end, we propose \textbf{Reflect-VLA}, a training framework that treats Reflective Guidance as privileged information and internalizes its planning benefits through \textbf{Dual-Path On-Policy Self-Distillation (Dual-OPSD)}. Starting from the same VLA checkpoint, the model supplied with Reflective Guidance acts as the frozen teacher, while the model receiving standard inputs acts as the student. The student first generates trajectories from the input available at deployment. A planning score then routes each trajectory to one of two learning paths. For risky or suboptimal trajectories, the teacher provides dense token-level distributions that encode corrective planning preferences. For reliable trajectories, a self-imitation objective preserves the student's existing decisions. The scalar score is therefore used only for trajectory-quality-aware routing, while the teacher distribution supplies the fine-grained supervision required for correction.

Effective self-distillation requires a teacher that can refine trajectories drawn from the student's rollout distribution. We construct this teacher using a progressive strategy comprising Driving-Knowledge Pretraining, Planning-and-Refinement SFT, and Self-Refinement Rejection Sampling. The first two stages establish driving knowledge, direct trajectory generation, and guidance-assisted refinement. The final stage exposes the teacher to student-generated failures and retains verified refinements, reducing the distribution gap between offline refinement data and on-policy trajectories. After Dual-OPSD, Reflect-VLA performs single-pass trajectory generation without Reflective Guidance, an online planning evaluator, or an iterative correction procedure.

Our main contributions are summarized as follows:
\begin{itemize}[leftmargin=*]
    \item We construct Reflective Guidance that converts trajectory evaluation results into structured, trajectory-specific diagnoses and corrective suggestions, and use it as privileged information available only during training.

    \item We propose Dual-OPSD, a trajectory-quality-aware self-distillation method that transfers corrective distributions for risky trajectories while preserving reliable planning.

    \item We develop a progressive teacher-construction strategy that combines driving-knowledge pretraining, planning-and-refinement supervision, and self-refinement rejection sampling to support stable guidance internalization and efficient single-pass planning.
\end{itemize}

%% file: sec/2_related.tex
\section{Related Work}

\subsection{VLA models for autonomous driving}
VLMs and VLA models have been increasingly studied for autonomous driving, with applications ranging from scene understanding~\cite{jiang2024senna,jiang2025alphadrive,marcu2024lingoqa} to direct trajectory prediction~\cite{xing2025openemma,zhao2025sce2drivex,liu2025reasonplan,fu2025orion}. To improve interpretability and planning accuracy, several methods introduce explicit intermediate reasoning, such as Chain-of-Thought reasoning~\cite{liu2025reasonplan,zhao2025sce2drivex,mtrdrive}. Their training pipelines typically begin with supervised fine-tuning on driving data and subsequently apply reinforcement learning, including Group Relative Policy Optimization, to improve planning performance~\cite{adathinkdrive,zhou2025autovla}. Although scalar planning rewards provide a direct measure of trajectory quality, they offer limited supervision when diverse failure causes receive similarly low scores~\cite{chen2026drivereward}. This issue is especially pronounced in long-tail scenarios, where repeated low-reward rollouts can make policy optimization inefficient. ELF-VLA~\cite{luo2026unleashing} addresses this limitation by using teacher-generated feedback to diagnose planning failures and construct improved training samples. Our work instead focuses on internalizing the benefit of such structured guidance so that it is no longer required during inference.

\subsection{On-Policy Distillation and On-Policy Self-Distillation}
On-policy distillation (OPD) and on-policy self-distillation (OPSD) are post-training approaches that supervise a student on trajectories sampled from its own policy~\cite{agarwal2024onpolicy,opsd}. Unlike off-policy distillation, this formulation queries a teacher with student-generated prefixes, thereby providing token-level supervision over the distribution encountered during training. OPD commonly relies on a separately trained teacher, whereas OPSD can derive the teacher from the same model by conditioning it on privileged information~\cite{ye2026onpolicycontextdistill,sang2026crisp}, such as ground-truth answers, additional instructions, or preference information. Recent work has combined dense token-level signals from OPD with reinforcement learning for VLA post-training~\cite{zhong2026vlaopd}, but its application to autonomous-driving planning remains limited. We formulate Reflective Guidance as privileged information and use it to obtain corrective teacher distributions on student-generated trajectories. Dual-OPSD further accounts for trajectory quality by distilling corrections only when needed and preserving decisions that are already reliable.

\begin{figure*}[htb]
    \centering
    \includegraphics[width=\textwidth]{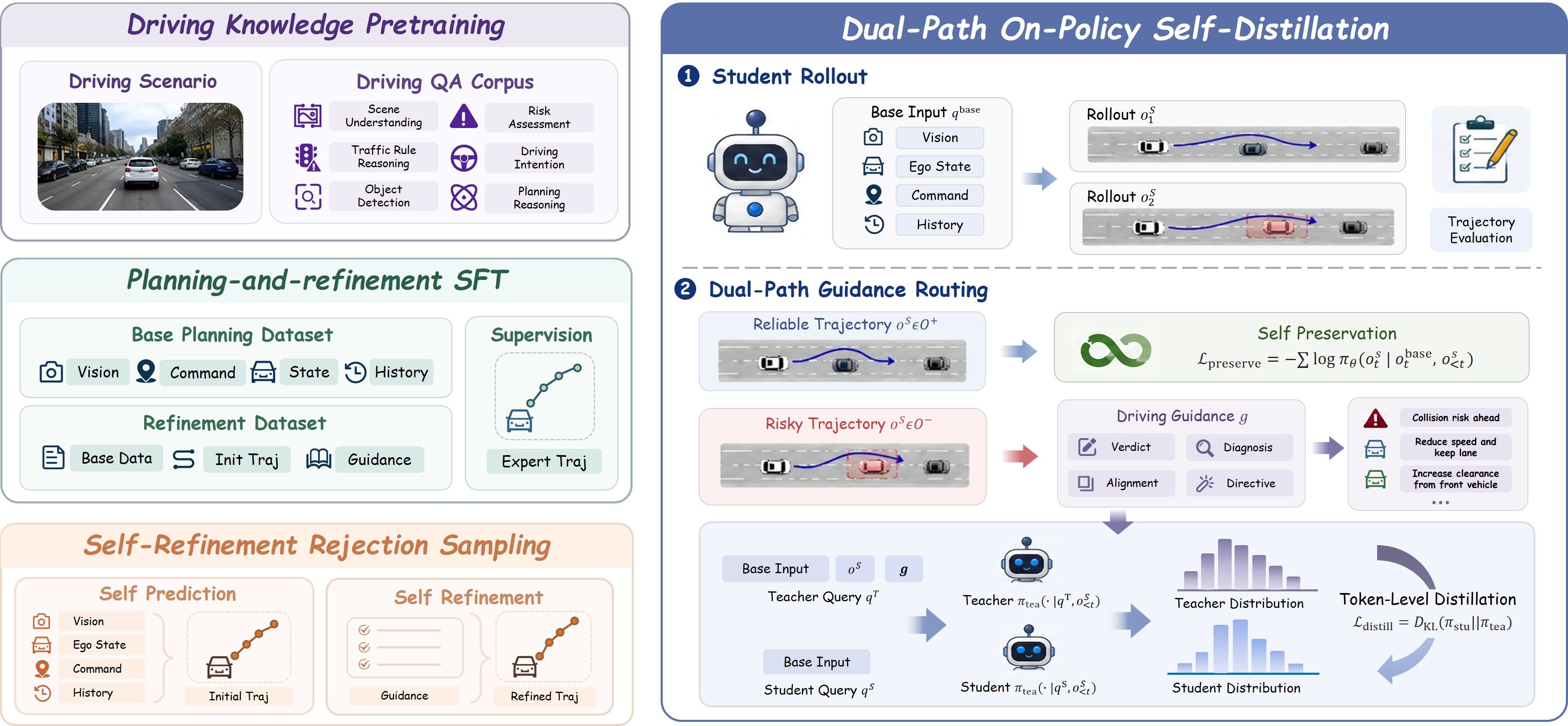}
    \caption{\textbf{Overview of \method.} Reflective Guidance is constructed from trajectory evaluation and used as privileged information to train a refinement-capable teacher. Dual-OPSD then transfers corrective distributions to the student on risky trajectories while preserving reliable student trajectories. Only the student is used for single-pass inference.}    
        \label{fig:framework}
\end{figure*}

%% file: sec/3_method.tex
\section{Method}

% Figure~\ref{fig:framework} illustrates the overall framework of our \method, comprising a three-stage training precess equipping the model with trajectory predicition and refinement ability, and the designed Dual-Path Guidance Online Policy Self-Distillation.
Figure~\ref{fig:framework} illustrates the overall framework of our \method, which comprises a three-stage training pipeline that equips the model with trajectory prediction and refinement capabilities, followed by the proposed Dual-Path On-Policy Self-Distillation.

\subsection{VLA Inputs Formulation}

We formulate end-to-end driving planning as conditional autoregressive trajectory generation. The standard driving input is
\begin{equation}
q^{\mathrm{base}}=
\left(
I_{\mathrm{cam}},
q_{\mathrm{cmd}},
q_{\mathrm{ego}},
q_{\mathrm{his}}
\right),
\end{equation}
where $I_{\mathrm{cam}}$ denotes the front-view camera observation, $q_{\mathrm{cmd}}$ is the high-level navigation command~(e.g., \textit{Go Straight}, \textit{Turn Left}, and \textit{Turn Right}), $q_{\mathrm{ego}}$ contains the ego-vehicle state~(e.g., velocity and acceleration), and $q_{\mathrm{his}}$ represents the historical ego trajectory. Given $q^{\mathrm{base}}$, the VLA autoregressively predicts future waypoints:
\begin{equation}
o=\{(x_t,y_t,h_t)\}_{t=1}^{T}
\sim
\pi_{\theta}(o\mid q^{\mathrm{base}}),
\end{equation}
where $(x_t,y_t,h_t)$ denotes the position and heading at the $t$-th future timestep. To cast continuous trajectory prediction as autoregressive token generation,
we convert the absolute waypoints into relative displacements $\{(\Delta x_t,\Delta y_t,\Delta h_t)\}_{t=1}^{T}$
and discretize these relative displacements into special trajectory tokens.
Details are provided in Appendix~\ref{appendix:data_prepare}.

To improve the planner's reasoning ability during training, we augment the base input with Reflective Guidance, forming the guidance-conditioned input
\begin{equation}
q^{\text{gui}}
=\left(q^{\text{base}},o^{\text{init}},g\right),
\end{equation}
where $o^{\text{init}}$ is the low-quality initial trajectory predicted under $q^{\text{base}}$, and $g$ denotes the designed \textbf{Reflective Guidance Corpus}, which comprises four components: (1) Trajectory Verdict, summarizing the overall trajectory evaluation; (2) Planning Diagnosis, identifying the dominant causes of planning failures; (3) Expert Alignment, relating these failures to discrepancies from expert trajectory; and (4) Refinement Directive, suggesting high-level corrective actions. The Reflective Guidance is designed to expose the reasoning process behind trajectory refinement, thereby guiding the model to generate safer and higher-quality trajectories. Details of the guidance construction are provided in Appendix~\ref{appendix:data_prepare}.

% During training, the planner receives either \(q\in
% \left\{
% q^{\text{base}},
% q^{\text{gui}}
% \right\}\).
% The base branch enables the model to learn direct trajectory generation from standard driving inputs, while the guidance-conditioned branch leverages guidance to acquire reflective correction capability and serves as a teacher that provides dense supervisory signals on the base policy's rollout trajectories during the OPSD stage.

% Our goal is to distill the planning capability acquired from the guidance-conditioned policy into the base policy through OPSD, enabling the model to retain strong planning and reasoning abilities even without guidence during inference:
% During training, the model receives either $q^{\text{base}}$ or $q^{\text{gui}}$. The base branch learns direct trajectory generation, whereas the guidance-conditioned branch learns reflective correction and later serves as the OPSD teacher. Our goal is to distill the planning capability of the guidance-conditioned teacher into the base student through OPSD, enabling the model to retain strong planning and reasoning abilities even without guidence during inference:

During training, the model is conditioned on either \(q^{\text{base}}\) or \(q^{\text{gui}}\). With \(q^{\text{base}}\), it learns direct trajectory generation, whereas the additional reflective guidance in \(q^{\text{gui}}\) enables trajectory  diagnosis and correction. Our goal is to distill the planning capability elicited by the guidance-conditioned input into direct trajectory generation through OPSD, enabling the model to retain strong planning and reasoning abilities even without guidance during inference:
\begin{equation}
o^{*}
=
\arg\max_{o}
\mathcal{P}(o\mid q^{\text{base}}).
\end{equation}

\subsection{Data Construction}

To develop both direct trajectory prediction and guidance-conditioned trajectory refinement capabilities, we first prepare the corresponding training data as follows.

\paragraph{Pretraining and Base Data.}

We first collect a diverse set of open-source driving QA datasets, including LingoQA~\cite{marcu2024lingoqa}, DriveGPT4~\cite{DriveGPT4}, Senna~\cite{jiang2024senna}, DriveLM~\cite{sima2024drivelm}, NuScenes-QA~\cite{qian2024nuscenes}, NuInstruct~\cite{ding2024holistic}, and DriveLMM~\cite{ishaq2025drivelmm}. Pretraining on these datasets equips the general VLM with domain-specific knowledge for autonomous driving, including road semantics, traffic regulations, and driving-intent understanding.

We then construct the base planning dataset, with each sample containing a front-view camera image, a high-level navigation command, the ego-vehicle state, a historical trajectory, and a ground-truth future trajectory represented as discretized special trajectory tokens.

\paragraph{Guidance-conditioned Refinement Data.}
% The effectiveness of Reflect-VLA relies on the availability of high-quality reflective guidance during training. Rather than providing only scalar planning scores, we construct a \textbf{Reflective Guidance Corpus} that transforms trajectory evaluation results into structured, trajectory-specific corrective supervision. 
% The objective of reflective guidance is not only to identify whether an initial trajectory is problematic, but also to explicitly describe \emph{why} it fails and \emph{how} to correct it.
To enable the model to refine low-quality initial trajectories, we further construct a guidance-conditioned refinement dataset. Specifically, for each driving scenario, we first build a candidate trajectory set
\(\mathcal{O}_{\text{can}}=\mathcal{O}_{\text{plan}}\cup\mathcal{O}_{\text{rule}}\),
where \(\mathcal{O}_{\text{plan}}\) denotes the trajectories generated by a classical planner, while \(\mathcal{O}_{\text{rule}}\) consists of trajectories obtained through rule-based perturbation.
Each candidate is assessed using the NAVSIM evaluator~\cite{dauner2024navsim}, together with map constraints and trajectory-level analysis. Based on the resulting scores, low-quality candidates are selected as the initial trajectories \(o^{\text{init}}\) for refinement. In particular, the NAVSIM evaluator provides the Predictive Driver Model Score (PDMS), which comprehensively measures collision safety, drivable-area compliance, time-to-collision, ego progress, and overall trajectory quality.
The Reflective Guidance $g$ is derived from the evaluation results, transforming scalar scores into structured, trajectory-specific corrective supervision. Consequently, training samples in the guidance-conditioned refinement dataset can be represented as:
\begin{equation}
\mathcal{D}_{\text{refine}} =\{ (q^{\text{gui}},o^{\text{ref}}) \}
=\{ (q^{\text{base}},o^{\text{init}},g,o^{\text{ref}}) \}.
\end{equation}
where $o^{\text{ref}}$ denotes the target refined trajectory, which can either be the ground-truth trajectory or a high-quality refined trajectory verified by the trajectory evaluator. 

% Based on these, we construct a structured Reflective Guidance $g$ comprising four components: (1) Overall Assessment, summarizing the overall trajectory evaluation; (2) Failure Analysis, identifying the dominant causes of planning failures; (3) Expert Comparison, relating these failures to discrepancies from expert trajectory; and (4) Actionable Correction, suggesting high-level corrective actions. Instead of providing low-level supervision, the Reflective Guidance is designed to expose the reasoning process behind trajectory refinement, thereby guiding the model to generate safer and higher-quality trajectories.
% \begin{equation}
% g=(g^{\text{eval}},g^{\text{reflect}},g^{\text{comparison}},g^{\text{action}}),
% \end{equation}
% where $g^{\text{eval}}$ summarizes the planning metrics, $g^{\text{reflect}}$ analyzes the primary planning deficiencies, $g^{\text{comp}}$ compares the initial trajectory with the expert trajectory, and $g^{\text{action}}$ provides actionable refinement instructions. Rather than providing low-level supervision, reflective guidance is designed to expose the reasoning process behind trajectory refinement. Specifically, it summarizes evaluation outcomes, identifies the dominant causes of planning failures, relates these failures to the discrepancies from expert behavior, and finally suggests high-level corrective actions. 
\paragraph{Self-Refinement Rejection Sampling.}
Training on the constructed planning and refinement data enables the model to leverage guidance for trajectory correction. However, because the initial trajectories in this dataset are generated offline, they may not fully capture the failure patterns encountered during model rollout. To mitigate this distribution mismatch, we introduce Self-Refinement Rejection Sampling. Specifically, the model first predicts an initial trajectory from the base input. When the trajectory is identified as problematic, the same model subsequently refines its prediction in response to the corresponding guidance. We retain only refined trajectories that satisfy the validity constraints and outperform their initial counterparts, and use these high-quality samples to further optimize the model.

\subsection{Three-stage SFT for Planning and Refinement}

% We propose a progressive training strategy comprising three stages: driving-knowledge pretraining, planning-and-refinement supervision fine-tuning, and self-refinement rejection sampling. 
% During pretraining, the model acquires essential driving knowledge. We then perform supervised fine-tuning using a mixture of base planning data and guidance-conditioned refinement data to teach the model both direct trajectory generation and guidance-conditioned trajectory refinement. Finally, rejection sampling produces a more reliable teacher for the subsequent Dual-OPSD stage by exposing the guidance-conditioned policy to realistic failures of the base policy.

We adopt a progressive three-stage training strategy consisting of driving-knowledge pretraining, planning-and-refinement supervised fine-tuning, and self-refinement rejection sampling. In the first stage, the model acquires essential domain knowledge for autonomous driving. In the second stage, it is trained on a mixture of base planning data and guidance-conditioned refinement data, enabling both direct trajectory generation and trajectory refinement. In the final stage, Self-refinement rejection sampling trains the model to correct failures from its own unguided rollouts, thereby yielding a more reliable teacher for subsequent on-policy self-distillation.

\subsection{Dual-Path On-Policy Self-Distillation}

% After the three-stage training process, the model acquires the ability to refine trajectories with the assistance of reflective guidance. Inspired by on-policy self-distillation in the LLM literature~\cite{opsd} where the same model acts as both teacher and student under different contexts and exploits privileged reasoning traces to teach its weaker counterpart, we propose \textbf{Dual-Path Guidance Online Policy Self-Distillation (Dual-OPSD)}, which transfers the planning capability learned under guidance conditions into the base policy.
After the three-stage training process, the model acquires the ability to refine trajectories with reflective guidance. Inspired by on-policy self-distillation for LLMs~\cite{opsd}, where a single model serves as both teacher and student under different input contexts, we propose \textbf{Dual-Path Online Policy Self-Distillation (Dual-OPSD)}, which distills the planning capability elicited by guidance-conditioned inputs into the model's direct trajectory generation behavior.

% In Dual-OPSD, the base policy serves as the \textit{student}, while the frozen guidance-conditioned policy serves as the \textit{teacher}. The student receives only the base input $q^{\text{base}}$ and performs on-policy rollout:
In Dual-OPSD, the same model serves as both the \textit{student} and the \textit{teacher} under different input contexts. With only the base input \(q^{\text{base}}\), it acts as the student and performs on-policy rollout:
\begin{equation}
o^{S}\sim\pi_{\theta}(\cdot|q^{\text{base}}),
\end{equation}
The generated trajectory is then evaluated by the NAVSIM evaluator. According to the resulting PDMS $r$, the rollout samples are divided into reliable trajectories $\mathcal{O}^{+}$ and risky trajectories $\mathcal{O}^{-}$ using a predefined threshold $s$:
\begin{equation}
o^{S}\in
\begin{cases}
\mathcal{O}^{+}, & r\geq s,\\
\mathcal{O}^{-}, & r<s,
\end{cases}
\end{equation}

This categorization enables two complementary optimization paths. For reliable trajectories in $\mathcal{O}^{+}$, we adopt a self-preservation objective to maintain the student's existing planning capability:
\begin{equation}
\mathcal{L}_{\text{preserve}}
=
-\sum_t
\log
\pi_{\theta}(o^{S}_t|q^{\text{base}},o^{S}_{<t}),
\end{equation}
where the generated trajectory serves as the self-supervised target. This prevents unnecessary modification of already valid behaviors during distillation.

For risky trajectories in $\mathcal{O}^{-}$, we construct the corresponding guidance $g$ based on the student rollout and evaluation results. A frozen version of the same model serves as the teacher and receives the privileged input
\(q^{T}=(q^{\text{base}},o^{S},g)\),
while the student remains restricted to $q^{\text{base}}$. The teacher's corrective planning preferences are transferred to the student by minimizing the token-level reverse KL divergence between their predictive distributions:
\begin{equation}
\mathcal{L}_{\text{distill}}
=
\sum_t
D_{\mathrm{KL}}
\left(
\pi_{\text{stu}}(\cdot|q^{\text{base}},o^{S}_{<t})
\|
\pi_{\text{tea}}(\cdot|q^{T},o^{S}_{<t})
\right),
\end{equation}

The overall Dual-OPSD objective combines the two pathways:
\begin{equation}
\mathcal{L}_{\text{Dual-OPSD}}
=
\mathbb{1}_{\{r\geq s\}} * \mathcal{L}_{\text{preserve}}
+
\mathbb{1}_{\{r<s\}} * \mathcal{L}_{\text{distill}}.
\end{equation}

By adaptively preserving reliable behaviors and distilling corrective preferences on failed rollouts, Dual-OPSD enables the model to progressively internalize the reflective planning capability acquired by the guidance-conditioned teacher while maintaining guidance-free inference.
\begin{table*}[t]
\centering
\renewcommand{\arraystretch}{1}
\begin{tabularx}{\textwidth}{l|c|c|XXXXX|c}
\toprule
Method & Image & Lidar & NC$\uparrow$ & DAC$\uparrow$ & TTC$\uparrow$ & CF$\uparrow$ & EP$\uparrow$ & PDMS$\uparrow$ \\
\midrule
% Constant Velocity & & & 68.0 & 57.8 & 50.0 & \textbf{100} & 19.4 & 20.6 \\
% Ego Status MLP & & & 93.0 & 77.3 & 83.6 & \textbf{100} & 62.8 & 65.6 \\
% \midrule
\multicolumn{9}{l}{\textbf{\textit{End-to-End Methods}}} \\
UniAD \cite{hu2023planning}
& \checkmark & & 97.8 & 91.9 & 92.9 & \textbf{100} & 78.8 & 83.4 \\

TransFuser \cite{chitta2022transfuser}
& \checkmark & \checkmark & 97.7 & 92.8 & 92.9 & \textbf{100} & 84.0 & 84.0 \\

DiffusionDrive \cite{liao2025diffusiondrive}
& \checkmark & \checkmark & 98.2 & 96.2 & 94.7 & \textbf{100} & 82.2 & 88.1 \\

WoTE \cite{li2025end}
& \checkmark & \checkmark & 98.5 & 96.8 & 94.9 & 99.9 & 81.9 & 88.3 \\

DVGT-2 \cite{zuo2026dvgt2}
& \checkmark & & 98.7 & 97.9 & 95.8 & \textbf{100} & 84.3 & 90.3 \\

\midrule
\multicolumn{9}{l}{\textbf{\textit{Vision-Language-Action Methods}}} \\
AutoVLA-3B \cite{zhou2025autovla}
& \checkmark & & 98.4 & 95.6 & \textbf{98.0} & 99.9 & 81.9 & 89.1 \\

AutoDrive-R$^2$-7B \cite{yuan2025autodrive}
& \checkmark & & 98.5 & 95.9 & 95.4 & \textbf{100} & 82.7 & 89.1 \\

DriveVLA-W0-7B \cite{li2025drivevla}
& \checkmark & & 98.7 & \textbf{99.1} & 94.9 & 99.3 & 83.3 & 90.2 \\

AdaThinkDrive-8B \cite{adathinkdrive}
& \checkmark & & 98.4 & 97.8 & 95.2 & \textbf{100} & 84.4 & 90.3 \\

ReCogDrive-2B \cite{li2025recogdrive}
& \checkmark & & 97.9 & 97.3 & 94.6 & \textbf{100} & \textbf{87.3} & 90.8 \\

ELF-VLA-8B \cite{luo2026unleashing}
& \checkmark & & 98.9 & 98.1 & 96.0 & \textbf{100} & 85.3 & 91.0 \\

% LaST-VLA-8B \cite{luo2026last}
% & \checkmark & & 98.7 & 97.9 & 95.6 & \textbf{100} & 86.8 & 91.3 \\

\midrule
Qwen3.5-0.8B~(RL)
& \checkmark & & 98.6 & 96.6 & 94.8 & \textbf{100} & 83.7 & 88.8 \\

Qwen3.5-2B~(RL)
& \checkmark & & 98.4 & 97.0 & 94.1 & \textbf{100} & 84.5 & 89.2 \\

Qwen3.5-4B~(RL)
& \checkmark & & \textbf{99.1} & 98.0 & 95.9 & \textbf{100} & 86.0 & 91.2 \\

\midrule
\textbf{\method-0.8B}
& \checkmark & & 98.8 & 96.7 & 95.8 & \textbf{100} & 83.2 & 89.3 \\

\textbf{\method-2B}
& \checkmark & & 98.9 & 97.0 & 96.3 & \textbf{100} & 84.2 & 90.0 \\

\textbf{\method-4B}
& \checkmark & & 98.9 & 98.0 & 96.7 & \textbf{100} & 86.0 & \textbf{91.5} \\
\bottomrule
\end{tabularx}
\caption{Comparison with state-of-the-art methods on NAVSIMv1 with PDMS.}
\label{table:main_table}
\end{table*}

\begin{table*}[t]
\centering
\resizebox{\textwidth}{!}{
\begin{tabular}{l|ccccc|cccc|c}
\toprule
Method & NC $\uparrow$ & DAC $\uparrow$ & DDC $\uparrow$ & TLC $\uparrow$ & EP $\uparrow$ & TTC $\uparrow$ & LK $\uparrow$ & HC $\uparrow$ & EC $\uparrow$ & EPDMS $\uparrow$ \\
\midrule
Ego Status 
& 93.1 & 77.9 & 92.7 & 99.6 & 86.0 & 91.5 & 89.4 & 98.3 & 85.4 & 64.0 \\

TransFuser~\cite{chitta2022transfuser} 
& 96.9 & 89.9 & 97.8 & 99.7 & 87.1 & 95.4 & 92.7 & 98.3 & 87.2 & 76.7 \\

DiffusionDrive~\cite{liao2025diffusiondrive} 
& 98.2 & 95.9 & 99.4 & \textbf{99.8} & 87.5 & 97.3 & 96.8 & 98.3 & \textbf{87.7} & 84.5 \\

HydraMDP++~\cite{li2024hydra} 
& 98.5 & 98.5 & 99.5 & 99.7 & 87.4 & 97.9 & 95.8 & 98.2 & 75.7 & 85.6 \\

DriveVLA-W0-7B~\cite{li2025drivevla} 
& 98.5 & \textbf{99.1} & 98.0 & 99.7 & 86.4 & 98.1 & 93.2 & 97.9 & 58.9 & 86.1 \\

% ELF-VLA-8B~\cite{luo2026unleashing} 
% & 99.0 & 98.2 & 99.4 & 99.7 & 89.5 & 98.3 & 97.5 & 98.4 & 86.9 & 87.3 \\
\midrule

% \rowcolor{gray!30}\textbf{\method-8B}~(SFT) 
% & 98.6 & 95.4 & 99.5 & 99.9 & 86.1 & 98.0 & 97.0 & 98.3 & 87.2 & 84.0 \\

\textbf{\method-0.8B} 
& 98.8 & 96.7 & 99.1 & 99.7 & 87.5 & 98.3 & 96.5 & 98.3 & 86.2 & 88.8 \\

\textbf{\method-2B} 
& \textbf{98.9} & 97.0 & 99.3 & 99.7 & 88.1 & 98.4 & 96.5 & \textbf{98.4} & 86.8 & 89.5 \\

\textbf{\method-4B} 
& \textbf{98.9} & 97.9 & \textbf{99.6} & \textbf{99.8} & \textbf{88.8} & \textbf{98.6} & \textbf{97.9} & 98.3 & 87.5 & \textbf{91.2} \\
\bottomrule
\end{tabular}
}
\caption{Comparison with state-of-the-art methods on NAVSIMv2 with EPDMS.}
\label{table:navsimv2}
\end{table*}

%% file: sec/4_exp.tex
\section{Experiment}

\subsection{Implementation details}

\textbf{Dataset.}
We evaluate trajectory generation on real-world driving scenarios using NAVSIM v1 and NAVSIM v2~\cite{dauner2024navsim,cao2025pseudo}, both of which provide simulation-based planning metrics. NAVSIM v2 further extends the scope of evaluation to traffic-rule compliance, lane-level behavior, and temporal consistency. To assess whether the learned driving capabilities generalize to interactive decision making, we additionally evaluate \method~on the closed-loop Bench2Drive benchmark~\cite{bench2drive}. Together, these benchmarks provide a comprehensive evaluation of both trajectory-level planning quality and closed-loop driving performance. Detailed dataset splits and evaluation settings are provided in Appendix~\ref{appendix:data_prepare}.

\paragraph{Metrics.}
To evaluate trajectory prediction and closed-loop driving performance, we employ benchmark-specific core metrics. We use the Predictive Driver Model Score (PDMS) for NAVSIM v1~\cite{dauner2024navsim} and the Extended Predictive Driver Model Score (EPDMS) for NAVSIM v2~\cite{cao2025pseudo}. For Bench2Drive, we report Driving Score (DS), Success Rate (SR), Efficiency (Effi), and Comfort (Comf). Detailed metric definitions are provided in Appendix~\ref{appendix:exp_details}.

\paragraph{Training Details.}
We use Qwen3.5-0.8B/2B/4B~\cite{qwen35blog} as the base model. Training consists of three stages: (1) driving-knowledge pretraining for 2 epochs with batch size 1 and a learning rate of \(1\times10^{-5}\); (2) fine-tuning on the planning, refinement, and rejection-sampling data for 2 epochs with batch size 4 and a learning rate of \(4\times10^{-5}\); and (3) on-policy self-distillation for 2 epochs with batch size 2 and learning rates of \(8\times10^{-6}\) and \(2\times10^{-6}\), respectively. Training is conducted on 8 NVIDIA H20 GPUs. Further details are provided in Appendix~\ref{appendix:exp_details}.

\subsection{Performance Comparison}

% bench2drive
% \begin{table*}[t!]
% \centering
% \vskip -0.5em
% \small
% \resizebox{\textwidth}{!}{
% \begin{tabular}{l|cccc}
% \toprule
% Method & Driving Score~(DS)$\uparrow$ & Success Rate~(SR)$\uparrow$ & Efficiency$\uparrow$ & Comfort$\uparrow$ \\
% \midrule
% % AD-MLP~\cite{zhai2305rethinking} & - & 18.1 & 0.0 & 48.5 & 22.6 \\
% UniAD-Base~\cite{hu2023planning} & 45.8 & 16.4 & 129.2 & 43.6 \\
% % VAD~\cite{jiang2023vad} & - & 42.4 & 15.0 & 157.9 & \textbf{46.0} \\
% DriveTransformer-L.~\cite{jia2025drivetransformer} & 63.5 & 35.0 & 100.6 & \textbf{46.0} \\
% DriveAdapter~\cite{jia2023driveadapter} & 64.2 & 33.1 & 70.2 & 16.0 \\
% ReasonPlan~\cite{liu2025reasonplan}  & 64.0 & 34.6 & \textbf{180.6} & 25.6 \\
% % MindDriver~\cite{zhang2026minddriver} & Multimodal CoT & 65.5 & 39.6 & 34.63 & 34.6 \\
% \midrule
% \textbf{\method-4B}  & \textbf{68.9} & \textbf{41.8} & 140.2  & 27.4 \\
% \bottomrule
% \end{tabular}
% }
% \caption{Comparison with state-of-the-art methods on Bench2Drive.}
% \label{table:b2d}
% \end{table*}

% Bench2Drive, single-column table
\begin{table}[t]
\centering
% \small
\renewcommand{\arraystretch}{1.2}
\resizebox{\columnwidth}{!}{
\begin{tabular}{l|cccc}
\toprule
Method
& DS$\uparrow$
& SR (\%)$\uparrow$
& Effi$\uparrow$
& Comf$\uparrow$ \\
\midrule
\multicolumn{5}{l}{\textbf{\textit{End-to-End Methods}}} \\
VAD~\cite{jiang2023vad}
& 42.35
& 15.00
& 157.94
& \textbf{46.01} \\

UniAD-Base~\cite{hu2023planning}
& 45.81
& 16.36
& 129.21
& 43.58 \\

DriveAdapter~\cite{jia2023driveadapter}
& 64.22
& 33.08
& 70.22
& 16.01 \\

\midrule
\multicolumn{5}{l}{\textbf{\textit{Vision-Language-Action Methods}}} \\

ReasonPlan~\cite{liu2025reasonplan}
& 64.01
& 34.55
& 180.64
& 25.63 \\

MindDriver~\cite{zhang2026minddriver}
& 65.48
& 39.55
& 143.21
& 34.63 \\

ORION~\cite{fu2025orion}
& 77.74
& 54.62
& 151.48
& 17.38 \\

AutoVLA~\cite{zhou2025autovla}
& 78.84
& 57.73
& 146.93
& 39.33 \\

\midrule

\textbf{\method-0.8B}
& 76.03
& 54.54
& 183.47
& 28.06 \\

\textbf{\method-2B}
& 77.39
& 55.90
& 174.15
& 27.89 \\

\textbf{\method-4B}
& \textbf{80.15}
& \textbf{59.09}
& \textbf{212.20}
& 21.92 \\

\bottomrule
\end{tabular}
}
\caption{Comparison with state-of-the-art methods on the Bench2Drive benchmark with DS (Driving Score), SR (Success Rate), Effi (Efficiency), and Comf (Comfortness).}
\label{table:b2d}
\end{table}

% student vs teacher vs opsd
\begin{table}[t]
\centering
\small 
\setlength{\tabcolsep}{4pt} 
\begin{tabular}{l|c|c|c}
\toprule
Method & Student & Teacher & OPSD\\
\midrule
\textbf{\method-0.8B} & 86.3 & 92.4 & 89.3\\
% \midrule
\textbf{\method-2B} & 87.4 & 92.7 & 90.0\\
% \midrule
\textbf{\method-4B} & 88.4 & 93.2 & 91.5\\
\bottomrule
\end{tabular}
\caption{Planning performance of the student, teacher, and student after Dual-OPSD on NAVSIMv1 using PDMS, where student and teacher refer to the same model after three-stage SFT performing trajectory prediction without and with guidance, respectively.}
\label{table:student vs teacher vs opsd}
\end{table}

\begin{table}[t]
\centering
\small 
\setlength{\tabcolsep}{4pt} 
\begin{tabular}{l|c|c|c}
\toprule
Base Model & SFT & GRPO & OPSD\\
\midrule
Qwen3.5-0.8B & 0.11 & 5.3 & 1.3 \\
% \midrule
Qwen3.5-2B & 0.26 & 22.3 & 1.7 \\
% \midrule
Qwen3.5-4B & 0.43 & 33.4 & 2.4 \\
\bottomrule
\end{tabular}
\caption{Comparison of training time (hours) per 10k samples for SFT, GRPO, and OPSD across different model scales.
}
\label{table:speed}
\end{table}

\paragraph{NAVSIM Benchmark.}
Table~\ref{table:main_table} and Table~\ref{table:navsimv2} report the planning performance on NAVSIM v1 and NAVSIM v2, respectively. On NAVSIM v1, Reflect-VLA consistently improves over the corresponding RL baselines across all model scales. Reflect-VLA-0.8B, 2B, and 4B achieve PDMS scores of 89.3, 90.0, and 91.5, outperforming their corresponding RL baselines by 0.5, 0.8, and 0.3 points, respectively. Notably, Reflect-VLA-4B achieves the best overall PDMS score, with improved TTC while maintaining strong performance on other metrics. These results validate the effectiveness and scalability of progressive guidance internalization: rather than relying solely on sparse scalar rewards, Reflect-VLA converts structured reflective guidance into fine-grained distributional supervision, enabling the policy to directly absorb trajectory-correction preferences. Together with its reduced training cost discussed below, Reflect-VLA provides a favorable balance between planning performance and optimization efficiency.

\paragraph{Bench2Drive Benchmark.}
Table~\ref{table:b2d} shows the closed-loop driving performance on the Bench2Drive benchmark. Our Reflect-VLA-4B achieves the best Driving Score (DS) of 80.15, Efficiency (Effi) of 212.20, and Success Rate (SR) of 59.09\%, outperforming existing baselines, including AutoVLA. This indicates that the guidance internalization possesses strong generalization and effectiveness in closed-loop evaluations involving complex interactions.

\paragraph{Effectiveness of Reflective Guidance.}

% Table~\ref{table:student vs teacher vs opsd} comparing the performance of the student, teacher, and student after OPSD. For both model scales, introducing the guidance consistently improves the planning performance, demonstrating that the reflective guidance enables the model to generate more reliable planning decisions. After OPSD, the distilled policy improves the base policy by 3.0 and 2.1 PDMS for the 0.8B and 4B models, respectively. Although a performance gap remains compared with guidance-conditioned policies, OPSD successfully internalizes substantial reasoning capability into the base policy, allowing \method~ to retain the benefits of reflective guidance while maintaining guidance-free inference.

Table~\ref{table:student vs teacher vs opsd} compares the performance of the student, the guidance-conditioned teacher, and the student after Dual-OPSD. Across all model scales, reflective guidance consistently improves planning performance, with the teacher outperforming the unguided student by 6.1, 5.3, and 4.8 PDMS for the 0.8B, 2B, and 4B models, respectively. After Dual-OPSD, the student gains 3.0, 2.6, and 3.1 PDMS over its original performance, substantially narrowing the gap to the teacher. These results demonstrate that Dual-OPSD effectively internalizes the planning knowledge elicited by reflective guidance, enabling \method~to retain most of its benefits while requiring no guidance during inference.

\paragraph{Training Efficiency.}

% Table~\ref{table:speed} presents the training time required by GRPO and OPSD for models of different scales. It shows that OPSD achieves substantially lower training costs across both model sizes compared with GRPO. Specifically, for the 0.8B model, OPSD reduces the training time per 10,000 samples from 5.3 hours to 1.3 hours, achieving approximately 4× higher efficiency. As the model scales to 4B parameters, the efficiency advantage becomes even more pronounced: OPSD requires only 2.4 hours, compared with 33.4 hours for GRPO, reducing the training cost by nearly 14×. Unlike GRPO, which requires expensive online exploration and repeated reward evaluation, OPSD directly transfers the guidance into dense token-level supervision, leading to more efficient optimization.
Table~\ref{table:speed} compares the training time of SFT, GRPO, and OPSD across different model scales. It shows that OPSD consistently requires less training time than GRPO. For the 0.8B model, OPSD reduces the training time per 10,000 samples from 5.3 to 1.3 hours, corresponding to a \(4.1\times\) speedup. This advantage increases with model scale: OPSD is \(13.1\times\) faster for the 2B model and \(13.9\times\) faster for the 4B model, requiring only 1.7 and 2.4 hours, respectively. Compared with GRPO, which requires expensive online exploration and iterative reward-based optimization, OPSD transfers guidance-conditioned knowledge through dense token-level supervision, enabling substantially more efficient training while maintaining competitive planning performance.

\subsection{Qualitative Analysis}

% Figure~\ref{fig:visualization} provides qualitative comparisons among the base policy, the guidance-conditioned policy, and the distilled policy different driving scenarios. 
% In the first scenario, the base policy generates a trajectory with potential collision risk. The guidance-conditioned policy identifies the safety issue and produces a more conservative longitudinal plan, which is subsequently internalized into the base policy through OPSD, enabling safer guidance-free inference. The token-level distributions further demonstrate the guidance effect: under the same rollout prefix, the guidance-conditioned policy prefers a more conservative action token (<|$x\_208$|>) than the base policy (<|$x\_215$|>). Through reverse KL minimization, OPSD aligns the base policy distribution with the guidance-conditioned policy, transferring the corresponding corrective behavior. The second scenario demonstrates a similar refinement process for drivable-area violations. The guidance-conditioned policy corrects the lateral deviation and generates a feasible trajectory, while OPSD transfers this capability into the base policy, enabling guidance-free generation of valid trajectories. More results are in Appendix~\ref{appendix:exp_results}.

\begin{figure*}[htb]
    \centering
    \includegraphics[width=\textwidth]{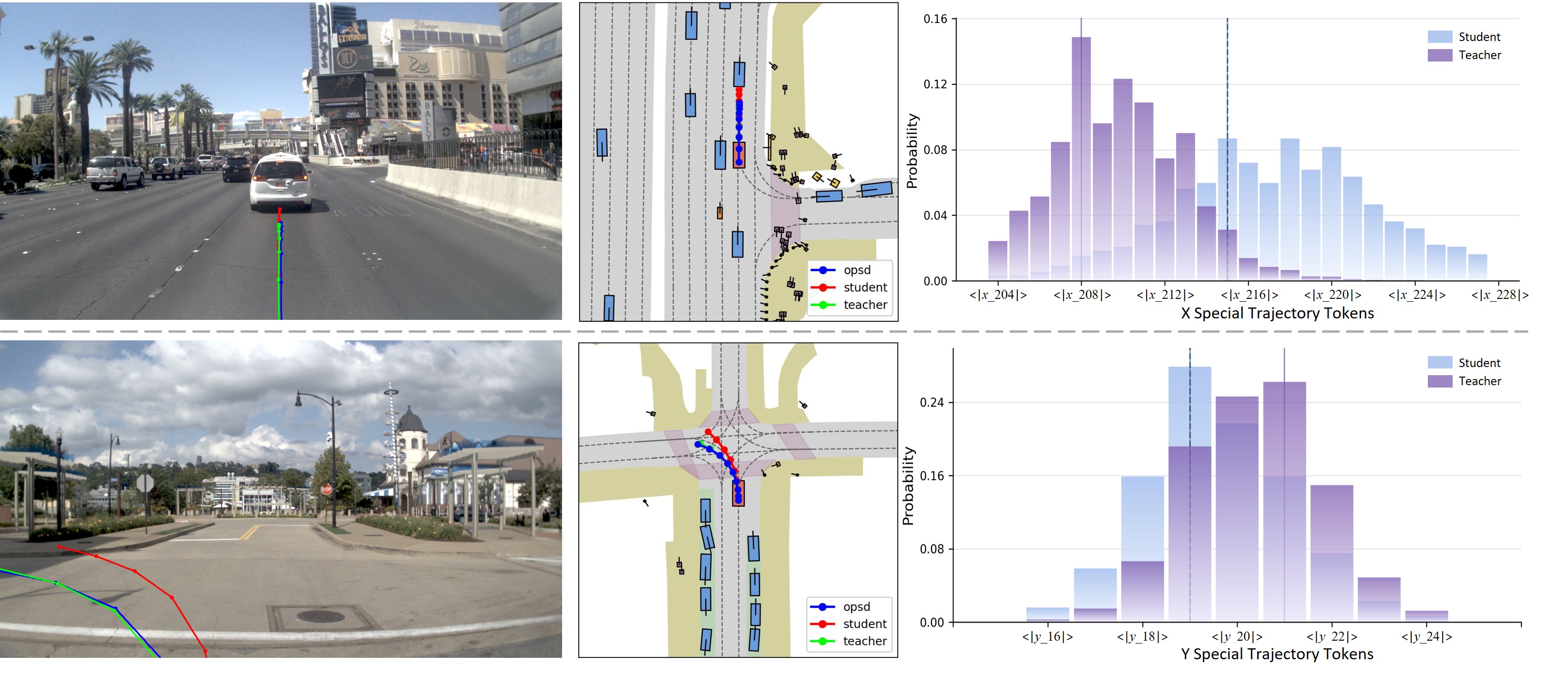}
    \caption{\textbf{Qualitative comparisons and token-level analysis.}
Left and middle: trajectories generated by the unguided student, guidance-conditioned teacher, and distilled student. Right: distributions over special trajectory tokens at critical decoding steps under the same student-generated prefix. Each token encodes a relative displacement quantized at 0.01\,m resolution; e.g., <|$x\_208$|> represents a 2.08\,m forward displacement for the next waypoint.}
    \label{fig:visualization}
\end{figure*}

Figure~\ref{fig:visualization} presents qualitative comparisons among the unguided student, the guidance-conditioned teacher, and the distilled student across representative driving scenarios. In the first scenario, the unguided student predicts a trajectory with potential collision risk, whereas the teacher produces a safer trajectory with reduced longitudinal progress. Dual-OPSD internalizes this corrective behavior into the student, enabling safer trajectory generation without reflective guidance at inference time. The token-level distributions provide further insight into this transfer. Given the same student-generated trajectory prefix, the teacher assigns the highest probability to the more conservative special trajectory token <|$x\_208$|>, while the student initially favors <|$x\_215$|>. Since the trajectory tokens encode relative displacements quantized at a resolution of 0.01\,m, these tokens correspond to forward displacements of 2.08\,m and 2.15\,m, respectively. By minimizing the reverse KL divergence, Dual-OPSD aligns the student distribution with the teacher distribution and transfers this fine-grained correction preference. The second scenario shows a similar effect for a drivable-area violation: the teacher corrects the lateral deviation, and the distilled student successfully reproduces a feasible trajectory without guidance. Additional qualitative results are provided in Appendix~\ref{appendix:vis}.

\begin{table}[t]
\centering
\small % 
% \begin{tabular}{l|ccccc|c}
\setlength{\tabcolsep}{5.0pt}
\resizebox{0.48\textwidth}{!}{
    \begin{tabular}{l|ccccc|c}
    \toprule
    Model & NC$\uparrow$ & DAC$\uparrow$ & TTC$\uparrow$ & CF$\uparrow$ & EP$\uparrow$ & PDMS$\uparrow$ \\
    \midrule
    SFT & 98.5 & 96.3 & 94.9 &100 & 81.7 & 88.1 \\
    Pre+SFT & 98.7 & 96.5 & 95.4 &100 & 82.1 & 88.5 \\
    Pre+SFT+RS & 98.6 & 96.4 & 95.3 &100 & 82.2 & 88.4 \\
    Pre+SFT+OPSD & 98.9 & 97.1 & 95.8 & 100& 84.7 & 90.2 \\
    Pre+SFT+RS+OPSD & \textbf{98.9} & \textbf{98.0} & \textbf{96.7} & \textbf{100} & \textbf{86.0} & \textbf{91.5}  \\
    \bottomrule
    \end{tabular}
}
\caption{Ablation study of training stages. We evaluate the effect of pretraining, supervised fine-tuning, rejection sampling and on-policy self-distillation on driving performance using NAVSIM evaluation metrics with the 4B model.}
\label{table:ablation_train}
\end{table}

%Ablation2
\begin{table}[h]
\centering
\small
\setlength{\tabcolsep}{3.3pt}

\begin{tabular}{c|cccc|c}
\toprule
\multirow{2}{*}{ID}
& Trajectory
& Planning
& Expert
& Refinement
& \multirow{2}{*}{PDMS$\uparrow$} \\
& Verdict
& Diagnosis
& Alignment
& Directive
& \\
\midrule
1 &  & \checkmark & \checkmark & \checkmark & 93.1 \\
2 & \checkmark &  & \checkmark & \checkmark & 92.4 \\
3 & \checkmark & \checkmark &  & \checkmark & 92.1 \\
4 & \checkmark & \checkmark & \checkmark &  & 92.9 \\
\midrule
\textbf{5}
& \checkmark
& \checkmark
& \checkmark
& \checkmark
& \textbf{93.2} \\
\bottomrule
\end{tabular}
\caption{Ablation study of reflective guidance components for the teacher model. We report the NAVSIM PDMS of the 4B teacher when conditioned on different combinations of guidance components.}
\label{table:fb_comp}
\end{table}

\subsection{Ablation Studies}

% \subsubsection{Ablation on Training Phase}
% Table~\ref{table:ablation_train} presents an ablation study of the training stages in Reflect-VLA. Pretraining improves PDMS from 88.0 to 88.5, showing the benefit of domain-specific initialization for learning fundamental driving knowledge. Self-Refinement Rejection Sampling further improves the final PDMS from 89.7 to 90.5, indicating that refinement samples generated from the model's own rollouts produce a more reliable teacher aligned with the trajectory distribution encountered during OPSD. Although these samples slightly reduce PDMS before OPSD, possibly due to forgetting basic planning abilities, OPSD effectively recovers this capability through guidance internalization.

% \subsubsection{Ablation on Reflective Guidance Component}
% Table~\ref{table:fb_comp} evaluates the impact of different guidance components. The results show that each component consistently improves the planning performance, confirming the effectiveness of our structured guidance design. In particular, \textit{Failure Analysis} and \textit{Expert Comparison} bring the largest gains, suggesting that providing explicit explanations of trajectory failures and their deviation from expert demonstrations is essential for learning effective refinement behaviors. 
%

\paragraph{Ablation on Training Stages.}

% Table~\ref{table:ablation_train} examines the contribution of each training stage in Reflect-VLA. Pretraining improves PDMS from 88.1 to 88.5, demonstrating the benefit of domain-specific initialization. Although adding Self-Refinement Rejection Sampling alone slightly decreases PDMS from 88.5 to 88.4, possibly due to partial forgetting of the base planning capability during further training on the refined samples, combining it with OPSD improves the final score from 89.7 to 90.5. This result indicates that refinement samples derived from the model's own rollouts better align the guidance-conditioned behavior with the failure distribution encountered during on-policy distillation, thereby providing more effective supervision for guidance internalization.
Table~\ref{table:ablation_train} examines the contribution of each training stage. Pretraining improves PDMS from 88.1 to 88.5, confirming the benefit of domain-specific initialization. Self-Refinement Rejection Sampling alone slightly reduces PDMS to 88.4, likely due to partial forgetting of base planning ability during further training on refined samples. However, combining it with OPSD raises the score from 90.2 to 91.5, indicating that refinement samples derived from the model's own rollouts better align the guidance-conditioned behavior with the failure distribution encountered during on-policy distillation, thereby providing more effective supervision for guidance internalization.

\paragraph{Ablation on Reflective Guidance Components.}

Table~\ref{table:fb_comp} evaluates the contribution of each reflective guidance component. The complete design achieves the best PDMS of 93.2, while removing any component degrades performance, confirming the effectiveness of our structured guidance design. In particular, \textit{Expert Alignment} and \textit{Planning Diagnosis} yield the largest gains, highlighting the importance of identifying planning failures and relating them to deviations from expert behavior.

%% file: sec/5_conclusion.tex
\section{Conclusion}
% In this paper, we proposed \method, a novel training framework that internalizes reflective guidance into VLA models for autonomous driving. By introducing a dual-path on-policy self-distillation method that distills token-level corrective distributions from the guidance-conditioned teacher to the base policy, our approach effectively transforms the benefits brought by the guidance into intrinsic planning capabilities, eliminating the need for explicit guidance or iterative refinement during inference. Experiments on NAVSIM, Bench2Drive, and nuScenes demonstrate that \method~ achieves competitive driving performance while offering a significant efficiency advantage, delivering up to a 10× training speedup over reinforcement learning baselines.

% In this paper, we introduce \method, a training framework that internalizes reflective guidance into VLA models for autonomous driving. Through dual-path on-policy self-distillation, \method~transfers token-level corrective distributions from guidance-conditioned generation to direct trajectory prediction, enabling strong planning without explicit guidance or iterative refinement at inference time. Experiments on NAVSIM, Bench2Drive, and nuScenes demonstrate competitive driving performance and up to a \(10\times\) training speedup over reinforcement-learning baselines.
In this paper, we introduce \method, a training framework that internalizes reflective guidance into VLA models for autonomous driving. Through dual-path on-policy self-distillation, \method~transfers token-level corrective distributions from guidance-conditioned generation to direct trajectory prediction, enabling strong planning without explicit guidance or iterative refinement at inference time. Experiments on NAVSIM and Bench2Drive demonstrate competitive driving performance and up to a \(10\times\) training speedup over reinforcement-learning baselines, while retaining efficient, guidance-free single-pass planning across different model scales.

%% file: sec/appendix.tex
\appendix
\section{Appendix} \label{appendix:appendix} 
In this appendix, we provide additional experimental results, data
construction details, implementation settings, and qualitative analyses to
complement the main paper. 
\textbf{Section~\ref{appendix:question}} addresses several potential questions about the proposed framework.
\textbf{Section~\ref{appendix:exp_results}} reports
the open-loop planning results on nuScenes and ablations on the top-$k$
distillation support, KL-divergence direction, and trajectory-routing
threshold. 
\textbf{Section~\ref{appendix:data_prepare}} details trajectory
tokenization, Reflective Guidance construction, pretraining data, and the
construction of planning, refinement, and self-refinement rejection-sampling
data. \textbf{Section~\ref{appendix:exp_details}} describes the model
architecture, training parameters, hardware configuration, and
benchmark-specific evaluation metrics. Finally,
\textbf{Section~\ref{appendix:vis}} presents detailed qualitative examples and
token-level analyses of how Dual-OPSD transfers corrective planning
preferences from the guidance-conditioned teacher to the student.

\section{Question} \label{appendix:question}

\noindent\textbf{Q1: Why not directly use Reflective Guidance during inference?}

\noindent\textbf{A:}
Constructing Reflective Guidance requires an initial trajectory and its
evaluation results, which introduces an additional prediction--evaluation--
refinement loop. Reflect-VLA instead treats the guidance as training-time
privileged information and internalizes its corrective preferences, enabling
single-pass inference without an online evaluator.

\vspace{0.5em}

\noindent\textbf{Q2: Why does Dual-OPSD use separate learning paths for reliable
and risky trajectories?}

\noindent\textbf{A:}
Teacher correction is useful for risky trajectories but may unnecessarily
modify predictions that are already reliable. Dual-OPSD therefore distills
guidance-conditioned teacher distributions for risky rollouts while applying
self-preservation to reliable ones, balancing correction and behavior
retention.

\vspace{0.5em}

\noindent\textbf{Q3: Why is Self-Refinement Rejection Sampling needed before
Dual-OPSD?}

\noindent\textbf{A:}
Offline refinement data may not cover the failure patterns produced by the
model itself. Self-Refinement Rejection Sampling exposes the teacher to
model-generated failures and retains only evaluator-verified improvements,
making its corrective behavior better aligned with the student rollout
distribution.

\vspace{0.5em}

\noindent\textbf{Q4: Does Reflective Guidance require an additional VLM or
manual feedback?}

\noindent\textbf{A:}
No. Reflective Guidance is constructed through deterministic templates using
trajectory-evaluation results, geometric differences from a reference
trajectory, and the navigation command. It therefore provides structured
corrective supervision without introducing another VLM or requiring manual
annotation.

\section{Additional Results} \label{appendix:exp_results}
\subsection{Main Results}
\noindent{\textbf{nuScenes Benchmark.}}  \label{appendix:nuscenes_exp}
Table~\ref{tab:nuscenes} reports the open-loop planning results on nuScenes.
\method-4B achieves state-of-the-art performance among VLA-based methods,
with the lowest average L2 error of 0.41\,m and collision rate of 0.14\%.
These results indicate that our approach improves geometric trajectory
accuracy while maintaining strong planning safety. The smaller
\method-2B model also achieves competitive results, with an average L2 error
of 0.52\,m and collision rate of 0.20\%, outperforming OpenDriveVLA and
approaching FSDrive. Even the 0.8B variant remains comparable to AutoVLA,
demonstrating consistent effectiveness across different model scales.

\begin{table*}[h]
\centering
\resizebox{\textwidth}{!}{
\begin{tabular}{l|l|cccc|cccc} 
\toprule
\multirow{2}{*}{\textbf{Method}} 
& \multirow{2}{*}{\textbf{VLM Model}} 
& \multicolumn{4}{c|}{\textbf{L2 (m)} $\downarrow$} 
& \multicolumn{4}{c}{\textbf{Collision (\%)} $\downarrow$} \\
\cmidrule(lr){3-6} \cmidrule(lr){7-10} 
& & 1s & 2s & 3s & Avg. & 1s & 2s & 3s & Avg. \\
\midrule

\multicolumn{10}{l}{\textbf{\emph{Traditional End-to-end Methods}}} \\
UniAD*~\cite{hu2023planning}
& -
& 0.20 & 0.42 & 0.75 & 0.46
& 0.02 & 0.25 & 0.84 & 0.37 \\
\midrule

\multicolumn{10}{l}{\textbf{\emph{Vision-Language-Action Methods}}} \\
Doe-1~\cite{zheng2024doe}
& Lumina-mGPT-7B
& 0.50 & 1.18 & 2.11 & 1.26
& 0.04 & 0.37 & 1.19 & 0.53 \\

AutoVLA*~\cite{zhou2025autovla}
& Qwen2.5-VL-3B
& 0.28 & 0.66 & 1.16 & 0.70
& 0.14 & 0.25 & 0.53 & 0.31 \\

RDA-Driver*~\cite{huang2024making}
& LLaVA-7B
& 0.23 & 0.73 & 1.54 & 0.80
& \textbf{0.00} & 0.13 & 0.83 & 0.32 \\

OpenDriveVLA*~\cite{zhou2025opendrivevla}
& Qwen2.5VL-3B
& 0.19 & 0.58 & 1.24 & 0.67
& 0.02 & 0.18 & 0.70 & 0.30 \\

FSDrive*~\cite{zeng2025futuresightdrive}
& Qwen2-VL-2B
& 0.18 & 0.39 & 0.77 & 0.45
& \textbf{0.00} & 0.06 & 0.42 & 0.16 \\
\midrule

\textbf{\method*~(Ours)}
& Qwen3.5-0.8B
& 0.30 & 0.69 & 1.20 & 0.73
& 0.15 & 0.27 & 0.57 & 0.33 \\

\textbf{\method*~(Ours)}
& Qwen3.5-2B
& 0.19 & 0.46 & 0.90 & 0.52
& 0.01 & 0.11 & 0.48 & 0.20 \\

\textbf{\method*~(Ours)}
& Qwen3.5-4B
& \textbf{0.16} & \textbf{0.35} & \textbf{0.71} & \textbf{0.41}
& \textbf{0.00} & \textbf{0.05} & \textbf{0.37} & \textbf{0.14} \\
\bottomrule
\end{tabular}
}
\caption{Comparison with state-of-the-art methods on nuScenes. * indicates that ego status is used.}
\label{tab:nuscenes}
\end{table*}

\subsection{Ablation Studies}

\begin{table}[t]
\centering
\small
\setlength{\tabcolsep}{5.0pt}
\resizebox{0.48\textwidth}{!}{
    \begin{tabular}{l|ccccc|c}
    \toprule
    Top-K & NC$\uparrow$ & DAC$\uparrow$ & TTC$\uparrow$ & CF$\uparrow$ & EP$\uparrow$ & PDMS$\uparrow$ \\
    \midrule
    Top-16
    & 98.7
    & 97.5
    & 95.1
    & \textbf{100}
    & 85.4
    & 90.2 \\

    Top-64 (ours)
    & \textbf{98.9}
    & \textbf{98.0}
    & \textbf{96.7}
    & \textbf{100}
    & \textbf{86.0}
    & \textbf{91.5} \\

    Top-256
    & \textbf{98.9}
    & 97.9
    & 96.5
    & \textbf{100}
    & \textbf{86.0}
    & 91.4 \\

    Full Vocab
    & \textbf{98.9}
    & 97.9
    & \textbf{96.7}
    & \textbf{100}
    & \textbf{86.0}
    & \textbf{91.5} \\
    \bottomrule
    \end{tabular}
}
\caption{Ablation study of the top-$k$ truncation used for token-level distillation on NAVSIM with the 4B model.}
\label{table:ablation_topk}
\end{table}

\paragraph{Effect of Top-$k$ Distillation.}
Table~\ref{table:ablation_topk} evaluates the size of the teacher-distribution
support retained during token-level distillation. Increasing $k$ from 16 to 64
improves PDMS from 90.2 to 91.5, since the larger support preserves more
informative teacher preferences. Further increasing $k$ to 256 or the full
vocabulary yields only marginal gains, indicating diminishing returns once the
dominant teacher distribution is sufficiently covered. We therefore use
$k=64$ as the distillation support size.

\begin{table}[t]
\centering
\small % 
% \begin{tabular}{l|ccccc|c}
\setlength{\tabcolsep}{5.0pt}
\resizebox{0.48\textwidth}{!}{
    \begin{tabular}{l|ccccc|c}
    \toprule
    Direction & NC$\uparrow$ & DAC$\uparrow$ & TTC$\uparrow$ & CF$\uparrow$ & EP$\uparrow$ & PDMS$\uparrow$ \\
    \midrule
    Forward KL & \textbf{98.9}   & 97.8     &   95.9  &  \textbf{100} & 85.8 &91.0 \\
    Reverse KL(ours) & \textbf{98.9} & \textbf{98.0} & \textbf{96.7} & \textbf{100} & \textbf{86.0} & \textbf{91.5}  \\
    \bottomrule
    \end{tabular}
}
\caption{Ablation study of the KL-divergence direction used for token-level distillation on NAVSIM with the 4B model.}
\label{table:ablation_kldir}
\end{table}

\paragraph{Effect of KL-Divergence Direction.}
Table~\ref{table:ablation_kldir} shows that reverse KL improves PDMS from 91.0 to
91.5. Reverse KL encourages the student to concentrate on the teacher's
high-probability corrective tokens and suppress unlikely alternatives. In
contrast, forward KL favors broader distribution coverage, which may dilute
the teacher's dominant correction preferences. We therefore adopt reverse KL
for token-level distillation.

\begin{table}[t]
\centering
\small
\setlength{\tabcolsep}{5.0pt}
\resizebox{0.48\textwidth}{!}{
    \begin{tabular}{l|ccccc|c}
    \toprule
    Threshold $s$ & NC$\uparrow$ & DAC$\uparrow$ & TTC$\uparrow$ & CF$\uparrow$ & EP$\uparrow$ & PDMS$\uparrow$ \\
    \midrule
    0.65
    & \textbf{98.9}
    & 97.8
    & 96.0
    & \textbf{100}
    & 85.9
    & 91.1 \\

    0.75
    & \textbf{98.9}
    & 97.8
    & 96.4
    & \textbf{100}
    & \textbf{86.0}
    & 91.3 \\

    0.85 (ours)
    & \textbf{98.9}
    & \textbf{98.0}
    & \textbf{96.7}
    & \textbf{100}
    & \textbf{86.0}
    & \textbf{91.5} \\

    0.95
    & \textbf{98.9}
    & 97.8
    & 96.2
    & \textbf{100}
    & \textbf{86.0}
    & 91.2 \\
    \bottomrule
    \end{tabular}
}
\caption{Ablation study of the PDMS routing threshold $s$ in Dual-OPSD on NAVSIM with the 4B model.}
\label{table:ablation_threshold}
\end{table}

\paragraph{Effect of the Routing Threshold.}
Table~\ref{table:ablation_threshold} shows that increasing $s$ from 0.65 to 0.85
gradually improves PDMS, as more suboptimal trajectories are routed to
distillation and aligned with the teacher's corrective preferences. However,
a larger threshold of 0.95 also distills many already reliable trajectories,
where unnecessary teacher intervention may disrupt effective student
behaviors. We therefore set $s=0.85$ to balance correction and preservation.

\section{Data Preparation Details}
\label{appendix:data_prepare}

\subsection{Trajectory Tokenization}

The future trajectory is represented as
$o=\{(x_t,y_t,h_t)\}_{t=1}^{T}$, where $x_t$, $y_t$, and $h_t$ denote the
longitudinal position, lateral position, and heading, respectively. To support
autoregressive trajectory generation, we convert the continuous trajectory
into structured tokens. Specifically, we prepend the current ego pose
$(x_0,y_0,h_0)=(0,0,0)$ and represent the motion between consecutive
timesteps by its relative displacement:
\begin{equation}
    (\Delta x_t,\Delta y_t,\Delta h_t)
    =
    (x_t-x_{t-1},\,y_t-y_{t-1},\,h_t-h_{t-1}),
\end{equation}

Each displacement component $d\in\{x,y,h\}$ is clipped to a predefined range
and uniformly quantized with resolution $\delta=0.01$, corresponding to 0.01~m for
the longitudinal and lateral displacements and 0.01~rad for the heading
change:
\begin{equation}
    k_t^d
    =
    \operatorname{round}\!\left(
    \frac{\operatorname{clip}(\Delta d_t,l_d,u_d)}{s}
    \right),
\end{equation}
We use $[l_x,u_x]=[-1.5,15.0]$, $[l_y,u_y]=[-5.0,5.0]$, and
$[l_h,u_h]=[-0.5,0.5]$, which cover the relative-motion ranges observed in
the datasets. This gives $k_t^x\in[-150,1500]$,
$k_t^y\in[-500,500]$, and $k_t^h\in[-50,50]$, corresponding to
1,651, 1,001, and 101 discrete values for the longitudinal, lateral, and
heading components, respectively.

Each quantized index is mapped to a dedicated special token. Non-negative
indices are represented as <|$x\_k$|>, <|$y\_k$|>, and
<|$theta\_k$|>, while a negative index $-k$ uses an $m$
prefix, such as <|$x\_mk$|>. At timestep $t$, the three tokens are
concatenated in longitudinal--lateral--heading order:
$a_t=\tau_x(k_t^x)\tau_y(k_t^y)\tau_{\theta}(k_t^h)$.
The complete trajectory is represented by the token sequence
$[a_1,\ldots,a_T]$. For example,
<|$x\_208$|><|$y\_m4$|><|$theta\_1$|> represents the relative motion
$(2.08,-0.04,0.01)$ between two consecutive timesteps.

During inference, the generated indices are converted back into relative
displacements and accumulated from the current ego pose:
\begin{equation}
    (\hat{x}_t,\hat{y}_t,\hat{h}_t)
    =
    \sum_{i=1}^{t}
    s\,(k_i^x,k_i^y,k_i^h),
    \qquad t=1,\ldots,T.
\end{equation}
This allows the autoregressively generated token sequence to be directly
decoded into the predicted future trajectory.

\subsection{Reflective Guidance Construction}
\label{appendix:guidance}

We construct Reflective Guidance for low-quality trajectories using a
deterministic template-based procedure. It relies on training-time privileged
information, including trajectory-evaluation results, the navigation command,
and a reference trajectory $o^{\mathrm{ref}}$, without requiring an
additional vision-language model. Given an initial trajectory
$o^{\mathrm{init}}$, we decode it into continuous waypoints when necessary
and generate four components: \textbf{Trajectory Verdict},
\textbf{Planning Diagnosis}, \textbf{Expert Alignment}, and
\textbf{Refinement Directive}.

\paragraph{Trajectory Verdict.}
The Trajectory Verdict converts evaluator outputs into an explicit reliability
judgment. It identifies whether the initial prediction is problematic and
summarizes the available PDMS, NC, DAC, TTC, EP, and CF metrics, providing
quantitative context for subsequent correction.

\paragraph{Planning Diagnosis.}
The Planning Diagnosis identifies the dominant causes of failure, including
collision risk, drivable-area violation, insufficient TTC margin, inadequate
progress, and trajectory misalignment. Safety-critical failures are
prioritized over efficiency-related errors, and only the most relevant
diagnoses are retained. When event-level information is available, the first
collision or boundary violation is localized to the corresponding waypoint
and position.

\paragraph{Expert Alignment.}
The Expert Alignment compares $o^{\mathrm{init}}$ with
$o^{\mathrm{ref}}$ and translates their geometric differences into
qualitative adjustments. These adjustments indicate whether to preserve or
shift the lateral position, increase or reduce longitudinal progress, and
maintain or correct the final heading. The navigation command is included to
keep the correction consistent with the intended maneuver.

\paragraph{Refinement Directive.}
The Refinement Directive converts the diagnosis and alignment results into an
ordered correction plan. Collision and TTC failures first require creating a
safer gap, while drivable-area violations require moving the path back into
the valid lane corridor. The directive then specifies the required lateral,
longitudinal, and heading adjustments, while preserving smooth intermediate
waypoints and navigation intent. A primary focus, such as scene
interpretation, lateral control, or longitudinal control, is assigned based on
the dominant failure.

\paragraph{Construction Flow.}
We first normalize the trajectories and evaluator outputs and determine
whether the initial trajectory requires correction. For a problematic trajectory,
we extract the dominant failures, derive reference-relative adjustments, and
instantiate the four components in the above order. The resulting guidance
$g$ is combined with the base input and initial trajectory as
\begin{equation}
    q^{\mathrm{gui}}
    =
    \left(
    q^{\mathrm{base}},
    o^{\mathrm{init}},
    g
    \right).
\end{equation}
The expert or evaluator-verified trajectory is used as the refinement target,
transforming scalar evaluation results and geometric discrepancies into
structured, trajectory-specific corrective supervision.

\subsection{Details of Pretraining Data}
\label{appendix:pretraining_data}

We collect several open-source driving QA datasets for driving-knowledge
pretraining, including LingoQA, DriveGPT4, Senna, DriveLM, NuScenes-QA,
NuInstruct, and DriveLMM.

\noindent\textbf{LingoQA.}
LingoQA contains approximately 419.9K QA pairs from short driving video clips,
covering scene description, localization, and reasoning.

\noindent\textbf{DriveGPT4.}
DriveGPT4 provides visual instruction data from multi-frame driving videos,
covering action interpretation, driving reasoning, question answering, and
low-level control prediction.

\noindent\textbf{Senna.}
Senna contains planning-oriented QA data over multi-view driving scenes,
covering scene understanding, motion prediction, and high-level planning
decisions.

\noindent\textbf{DriveLM.}
DriveLM contains approximately 0.3M image-question pairs from 696 nuScenes
scenes for graph-based driving-scene reasoning.

\noindent\textbf{NuScenes-QA.}
NuScenes-QA provides approximately 460K QA pairs over multi-frame camera and
LiDAR observations.

\noindent\textbf{NuInstruct.}
NuInstruct contains approximately 91K multi-view video-QA pairs across 17
spatial, temporal, and multi-view reasoning tasks.

\noindent\textbf{DriveLMM.}
DriveLMM provides over 18K training VQA examples with step-by-step
reasoning for perception, prediction, and planning.

\subsection{Details of SFT Data}

\paragraph{Planning Data.}
For each dataset, we construct a base planning set in which each sample
contains a front-view camera image, the ego-vehicle state including velocity
and acceleration, a high-level navigation command (\emph{Go Straight},
\emph{Turn Left}, or \emph{Turn Right}), a historical ego trajectory, and the
ground-truth future trajectory. Both historical and future trajectories are
sampled at 0.5-second intervals, and the future trajectory is converted into
discrete special tokens for autoregressive prediction.

\paragraph{Guidance-Conditioned Refinement Data.}
Each refinement sample pairs the base driving input with a low-quality initial
trajectory, its corresponding Reflective Guidance, and a reference trajectory.
The initial trajectories are obtained from two complementary sources:
\begin{itemize}
    \item \textbf{Planner-generated trajectories.} We collect trajectories
    produced by a classical planner and retain those exhibiting correctable
    failures in safety, drivable-area compliance, or progress.
    \item \textbf{Rule-perturbed trajectories.} We perturb valid trajectories
    along the longitudinal, lateral, or heading dimensions to construct
    representative planning failures while preserving a valid trajectory
    format.
\end{itemize}
All candidates are evaluated using the trajectory evaluator and map
constraints, and low-quality but valid trajectories are selected as
$o^{\mathrm{init}}$. The ground-truth expert trajectory is used as the
refinement target; alternatively, an evaluator-verified refined trajectory is
retained when it resolves the failure and improves upon the initial
trajectory.

\paragraph{Self-Refinement Rejection Sampling Data.}
We construct self-refinement data from the model's own rollouts. For each base
input, the model first generates an unguided initial trajectory, which is
evaluated to identify problematic cases. We then construct the corresponding
Reflective Guidance and prompt the same model to generate a refined
trajectory. A sample is retained only when the refined trajectory satisfies
the trajectory and map constraints and achieves a higher evaluation score
than the initial trajectory. When multiple valid refinements are available,
we retain the highest-scoring one.
\section{Experimental Details}  \label{appendix:exp_details}
% \subsection{Implementation Details}

\subsection{Training Parameters and Hardware Configuration.} \label{appendix:param}
We train Qwen3.5-0.8B/2B/4B on 8 NVIDIA H20 GPUs. Driving-knowledge
pretraining is conducted for 2 epochs with a batch size of 1 and a learning
rate of $1\times10^{-5}$. Planning-and-Refinement SFT is then performed on
the base planning and guidance-conditioned refinement data, followed by
Self-Refinement Rejection Sampling on the retained self-refinement data.
These two stages use the same configuration of 2 epochs, a batch size of 4,
and a learning rate of $4\times10^{-5}$. Dual-OPSD is subsequently performed
for 2 epochs with a batch size of 2 and learning rates of $8\times10^{-6}$
and $2\times10^{-6}$ for the first and second epochs, respectively. During
Dual-OPSD, on-policy rollouts are generated through greedy decoding, with the
temperature set to 0. For token-level distillation, we retain the teacher's
top-$k$ support with $k=64$ and minimize the reverse KL divergence over the
truncated distribution. The routing threshold is set to $s=0.85$. Reliable
trajectories are optimized through self-preservation, whereas risky
trajectories receive token-level supervision from the frozen
guidance-conditioned teacher. Detailed hyperparameters are provided in
Table~\ref{tab:hyperparameters}.

% \begin{table}[h!]
% \centering
% \setlength{\tabcolsep}{8.0pt} 
% \caption{Training hyperparameters of \method~on NAVSIM benchmark.}
% \begin{tabular}{l|ll}
% \midrule
% \textbf{Stage} & \textbf{Hyper-parameter} & \textbf{Value} \\ 
% \midrule
% \multirow{10}{*}{SFT-Stage1} & Epochs & 2 \\
%  & Batch size & 2 \\ 
%  & Learning rate & \(1 \times 10^{-5}\) \\  
%  & Gradient accumulation steps & 2 \\ 
%  & Weight decay & 0.05 \\ 
%  & Weight ratio & 0.05 \\ 
%  & Weight $\lambda_{action}$ & 0.01 \\ 
%  & Weight $\lambda_{WM}$ & 1.00 \\ 
%  & Weight $\lambda_{3D}$ & 1.00 \\ 
%  & Structured Causal Mask & True \\
% \midrule
% \multirow{10}{*}{SFT-Stage2} & Epochs & 2 \\
%  & Batch size & 2 \\
%  & Learning rate & \(1 \times 10^{-5}\) \\
%  & Gradient accumulation steps & 2 \\ 
%  & Weight decay & 0.05 \\
%  & Weight ratio & 0.05 \\ 
%  & Weight $\lambda_{action}$ & 1.00 \\ 
%  & Weight $\lambda_{WM}$ & 0.01 \\ 
%  & Weight $\lambda_{3D}$ & 0.01 \\ 
%  & Structured Causal Mask & False \\
% \midrule
% \multirow{13}{*}{RL} & Epochs & 2 \\
%  & Batch size & 2 \\
%  & Learning rate & \(2 \times 10^{-6}\) \\ 
%  & Gradient accumulation steps & 16 \\ 
%  & Weight decay & 0.05 \\ 
%  & Weight ratio & 0.05 \\  
%  & Number of generations & 8 \\ 
%  & Number of iterations & 2 \\
%  & KL penalty $\beta$ & 0.1 \\
%  & Temperature & 2.0 \\ 
%  & Weight $\lambda_{traj}$ & 8 \\ 
%  & Weight $\lambda_{fmt}$ & 1 \\ 
%  & Weight $\lambda_{goal}$ & 1 \\ 
% \midrule
% \end{tabular}
% \label{tab:hyperparameters}
% \end{table}

\begin{table}[h!]
\centering
\setlength{\tabcolsep}{6.0pt}
\renewcommand{\arraystretch}{1.05}
\resizebox{\columnwidth}{!}{%
\begin{tabular}{l|ll}
\toprule
\textbf{Stage} & \textbf{Hyper-parameter} & \textbf{Value} \\
\midrule

\multirow{7}{*}{Pretraining}
 & Epochs & 2 \\
 & Batch size & 1 \\
 & Learning rate & \(1\times10^{-5}\) \\
 & Gradient accumulation steps & 2 \\
 & Optimizer & Adam \\
 & Weight decay & 0.05 \\
 & Warmup ratio & 0.1 \\
\midrule

\multirow{7}{*}{\shortstack{SFT \&\\ Rejection \\ Sampling}}
 & Epochs & 2 \\
 & Batch size & 4 \\
 & Learning rate & \(4\times10^{-5}\) \\
 & Gradient accumulation steps & 2 \\
 & Optimizer & Adam \\
 & Weight decay & 0.05 \\
 & Warmup ratio & 0.1 \\
\midrule

\multirow{14}{*}{Dual-OPSD}
 & Epochs & 2 \\
 & Batch size & 2 \\
 & Learning rate for epoch1 & \(8\times10^{-6}\) \\
 & Learning rate for epoch2 & \(2\times10^{-6}\)  \\
 & Gradient accumulation steps & 2 \\
 & Optimizer & Adam \\
 & Weight decay & 0.05 \\
 & Warmup ratio & 0.1 \\
 & Routing threshold \(s\) & 0.85 \\
 & Rollout temperature & 0.0 \\
 & Distillation support top-$k$ & 64 \\
 & Distillation objective & Reverse KL \\
 & Teacher update & Frozen \\
 & Rollouts per input & 1 \\

\bottomrule

\end{tabular}%
}
\caption{Training hyperparameters of \method.}
\label{tab:hyperparameters}
\end{table}

\subsection{Metrics.}
\label{appendix:metric}
We use benchmark-specific metrics to comprehensively evaluate planning
performance. For the NAVSIM suite, we report the Predictive Driver Model Score
(PDMS) on \textbf{NAVSIMv1}~\cite{dauner2024navsim} and the Extended
Predictive Driver Model Score (EPDMS) on
\textbf{NAVSIMv2}~\cite{cao2025pseudo}. For \textbf{Bench2Drive}, we evaluate
interactive closed-loop performance using Driving Score (DS), Success Rate
(SR), Efficiency (Effi), and Comfort (Comf). In addition, following
UniAD~\cite{hu2023planning}, we evaluate open-loop trajectory quality on
nuScenes using L2 error and collision rate.

For NAVSIMv1, PDMS combines five sub-metrics: No At-Fault Collision (NC),
Drivable Area Compliance (DAC), Time-to-Collision (TTC), Comfort (CF), and Ego
Progress (EP). It is computed as
\begin{equation}
PDMS = NC \times DAC \times \left( \frac{5\times EP + 5\times TTC + 2\times CF}{12} \right),
\end{equation}
where NC and DAC serve as multiplicative penalties, while EP, TTC, and CF
contribute to the weighted planning score.

For NAVSIMv2, EPDMS extends the evaluation with additional safety, compliance,
lane-level, and comfort metrics. It includes No At-Fault Collision (NC),
Drivable Area Compliance (DAC), Driving Direction Compliance (DDC), Traffic
Light Compliance (TLC), Ego Progress (EP), Time-to-Collision (TTC), Lane
Keeping (LK), History Comfort (HC), and Extended Comfort (EC). The score is
defined as
\begin{equation}
\begin{split}
    & EPDMS ={} NC \times DAC \times DDC \times TLC \times  \\& \left( \frac{5\times EP + 2\times LK + 2\times HC + 5\times TTC + 2\times EC}{16} \right),
\end{split}
\end{equation}

For Bench2Drive, DS combines route completion with multiplicative penalties
for driving infractions, while SR measures the proportion of routes completed
within the time limit without traffic violations:
\begin{equation}
\begin{aligned}
&\text{Success Rate} = \frac{n_{\text{success}}}{n_{\text{total}}}, \\
&\text{Driving Score} = \frac{1}{n_{\text{total}}} \sum_{i=1}^{n_{\text{total}}} \text{RC}_i \prod_{j=1}^{n_{i,\text{penalty}}} p_{i,j}.
\end{aligned}
\end{equation}
Here, $n_{\text{success}}$ and $n_{\text{total}}$ denote the numbers of
successful and evaluated routes, respectively; $\text{RC}_i$ is the route
completion of route $i$, and $p_{i,j}$ is its $j$-th infraction penalty.
Efficiency averages the ego-speed ratio relative to nearby vehicles, while
Comfort reports the proportion of trajectory segments satisfying predefined
kinematic smoothness thresholds.

For nuScenes, we report the L2 error between the planned and expert
trajectories together with the collision rate.

\section{Visualization Analysis} \label{appendix:vis} 
\begin{figure*}[htb]
    \centering
    \includegraphics[width=\textwidth]{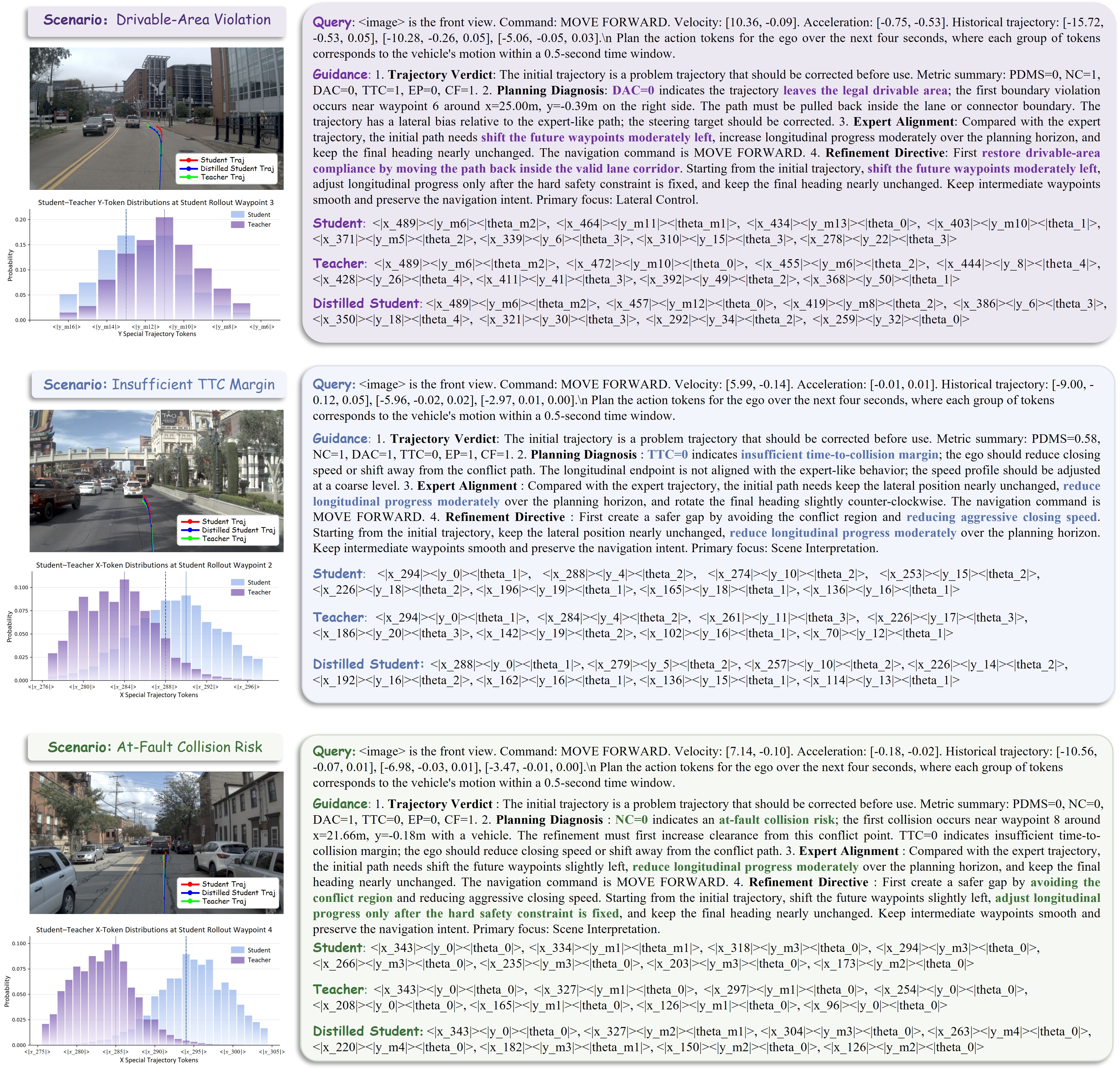}
    \caption{\textbf{Detailed qualitative visualization of Reflect-VLA.}
We present three representative failure cases: \emph{Drivable-Area Violation},
\emph{Insufficient TTC Margin}, and \emph{At-Fault Collision Risk}. For each
scenario, we show the front-view image, the initial student trajectory, the
guidance-conditioned teacher trajectory, and the distilled student trajectory,
together with the constructed Reflective Guidance and the corresponding
token-level distribution comparison at a critical waypoint under the same
student-generated prefix. The results illustrate how Reflective Guidance
shifts the teacher toward safer and more expert-aligned trajectory tokens, and
how Dual-OPSD transfers these corrective preferences to the distilled student.}
    \label{fig:appendix_visualization}
\end{figure*}

Figure~\ref{fig:appendix_visualization} presents three representative failure cases. In each scenario, the unguided student produces a problematic trajectory, while the guidance-conditioned teacher generates a safer correction. After Dual-OPSD, the distilled student largely reproduces this behavior without requiring guidance at inference time. For the \emph{Drivable-Area Violation}, the teacher shifts the trajectory away from the road boundary and back into the valid lane corridor. For the \emph{Insufficient TTC Margin}, it reduces forward progress to maintain a safer distance from the conflict region. For the \emph{At-Fault Collision Risk}, it increases clearance and postpones longitudinal progress until the hard safety constraint is resolved. The distilled student exhibits similar lateral and longitudinal adjustments in all three cases. The token-level distributions further reveal the transfer process. Under the same student-generated prefix, the teacher assigns greater probability to tokens corresponding to safer lateral or longitudinal motions, whereas the original student favors riskier alternatives. Dual-OPSD shifts the student toward these corrective preferences, resulting in safer token selection and improved trajectory generation

